 \documentclass[pmlr, twocolumn]{jmlr} 

\usepackage{booktabs}
\usepackage[load-configurations=version-1]{siunitx} 

\usepackage{pifont}
\newcommand{\xmark}{\ding{55}}

\usepackage{capt-of}

\theorembodyfont{\upshape}
\theoremheaderfont{\scshape}
\theorempostheader{:}
\theoremsep{\newline}

\title[Progress of machine learning for structured medical data]{Evaluating Progress on Machine Learning for Longitudinal Electronic Healthcare Data}

\author{%
\Name{David Bellamy} \Email{david\_bellamy@g.harvard.edu} \\
\Name{Andrew Beam} \Email{andrew\_beam@hms.harvard.edu}\\
\addr Department of Epidemiology \\Harvard T.H. Chan School 
 Public Health
\AND
\Name{Leo Celi} \Email{leoanthonyceli@yahoo.com}\\
\addr Massachusetts Institute of Technology \\
Beth Israel Deaconess Medical Center
}

\begin{document}

\maketitle
\begin{abstract}
The Large Scale Visual Recognition Challenge based on the well-known Imagenet dataset catalyzed an intense flurry of progress in computer vision. Benchmark tasks have propelled other sub-fields of machine learning forward at an equally impressive pace, but in healthcare it has primarily been image processing tasks, such as in dermatology and radiology, that have experienced similar benchmark-driven progress. In the present study, we performed a comprehensive review of benchmarks in medical machine learning for structured data, identifying one based on the Medical Information Mart for Intensive Care (MIMIC-III) that allows the first direct comparison of predictive performance and thus the evaluation of progress on four clinical prediction tasks: mortality, length of stay, phenotyping, and patient decompensation. We find that little meaningful progress has been made over a 3 year period on these tasks, despite significant community engagement. Through our meta-analysis, we find that the performance of deep recurrent models is only superior to logistic regression on certain tasks. We conclude with a synthesis of these results, possible explanations, and a list of desirable qualities for future benchmarks in medical machine learning.
\end{abstract}


\section{Introduction}
\label{introduction}

The collection and use of electronic health records (EHRs) has a long history, spanning nearly 50 years, starting with the development of systems like COSTAR, PROMIS, TMR, and HELP \citep{warner1983help, barnett1976computer, schultz1971sv, stead1988computer}. Despite initial barriers, the early 1990s witnessed the adoption of EHRs as they became integrated into the physician workstation \citep{litt1992digital, higgins1991graphical}, and entire hospital departments were interfaced with them. The hope of a learning healthcare system enabled by EHRs has been one of the grand challenges in medical informatics \citep{halamka2008early, mandl2012escaping}, and there is great enthusiasm recently that machine learning may finally unlock this potential \citep{beam2018big}.

In recent years, machine learning, and in particular deep learning, has experienced a surge in interest and has become viewed as the state of the art for pattern recognition. This reputation has grown out of the remarkable progress measured on benchmark datasets in areas such as computer vision and natural language processing. 

\begin{figure*}
  \label{fig:ImageNetprogress}
  \centering
  \includegraphics[width=0.8\linewidth]{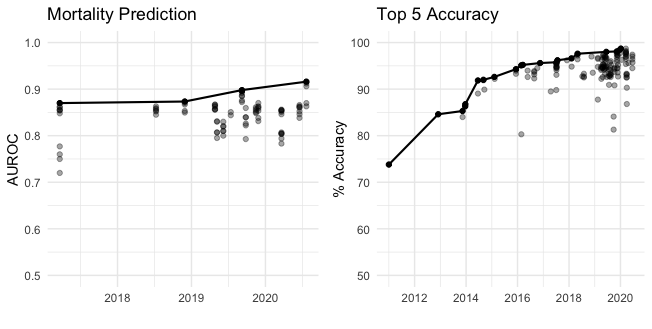}
  \caption{The progress of mortality prediction in the MIMIC-III dataset compared to image classification on ImageNet. The left graph shows the AUROC for all mortality prediction models developed on the benchmark established by \citet{harutyunyan2019multitask}, whereas the right graph shows the top-5 classification accuracy on ImageNet. The lines on each graph connect the performances of the best model over time.}
\end{figure*}

For example, in 2009 Deng et al. published the initial release of the ImageNet dataset, which included 3.2 million annotated images with greater diversity and accuracy than any other image dataset at that time \citep{deng2009imagenet}. The following year, the annual ImageNet Large Scale Visual Recognition Challenge (ILSVRC) was launched, motivating computer vision researchers to improve on existing models using the same experimental set-up. Over the last 10 years, top-1 and top-5 image classification accuracy on ImageNet have increased from 50.9\% to 88.5\%, and 73.8\% to 98.7\%, respectively \footnote{Data from: \url{https://paperswithcode.com/sota/image-classification-on-imagenet}}. However, progress on mortality prediction, a popular healthcare prediction task, has been far less significant, as can be seen in Figure \ref{fig:ImageNetprogress}.

Evaluating progress of machine learning for healthcare is easier done for medical image processing tasks, such as in dermatology and radiology, due to the lack of privacy concerns which enables easier sharing \citep{beam2016translating}. Benchmark datasets, like CheXpert \citep{DBLP:journals/corr/abs-1901-07031}, SD-198 \citep{sun2016benchmark}, and the annual International Skin Imaging Collaboration (ISIC) challenge dataset \citep{rotemberg2020patient} have been established in very much the same fashion as ImageNet, and the same prediction tasks (e.g. image classification, object localization) and evaluation metrics have been adopted. 

However, there is a lack of literature that measures the progress of modeling structured medical data, such as the data in EHRs, and it is not obvious that deep learning will be strictly better than traditional methods on this kind of `tabular` data relative to imaging or text data \citep{schmaltz2020sharpening, christodoulou2019systematic}. In order to have a similar discussion of progress, the performance of separately developed models must be directly comparable, requiring the usage of identical train/test datasets, as well as common prediction tasks and evaluation metrics. 

In this paper, we conducted a comprehensive review of the machine learning for healthcare literature using Web of Science, focusing on papers that construct benchmarks for predictions using structured medical data. To our knowledge, this is the first attempt to aggregate the evaluations of models developed on identical experimental set-ups in the context of structured medical data. In our review, we found one benchmark paper based on the MIMIC-III dataset \citep{johnson2016mimic} that has had 190 citations since it was first published in 2017, 19 of which develop and test a total of 178 models (including baselines) on the same experimental set-up \citep{harutyunyan2019multitask}. This group of 20 papers allows the first direct comparison of models developed for longitudinal clinical predictions using EHR data, and structured medical data more generally. Section \ref{results} quantifies the changes in model performance over time, by model type, and by evaluation metric. Our results show that progress in predictive performance on these tasks has been rather stagnant, and that deep learning models do not perform better than traditional approaches like logistic regression in most cases, raising questions about the choice of datasets, pre-processing steps, task definitions and evaluation metrics used in machine learning for structured healthcare data. 

Ultimately, we argue that the alignment of common goals and practices plays a central role in the progress of machine learning for healthcare, particularly for longitudinal predictions using structured medical data. We conclude with a thorough discussion of the pitfalls that limit this progress.

\section{Methods}
\label{methods}

\subsection{Study selection}
\label{study selection}

The core collection of the Web of Science was searched for the following query: ALL FIELDS: ((benchmark  OR benchmarking)  AND (medical  OR healthcare  OR medicine  OR clinical)  AND (machine learning)). The earliest publication matching this query was published in 1995, so our search spanned from 1995 to current day. These publications were reviewed for any that were proposing a benchmark experimental set-up for a clinical prediction task using structured data. We define a benchmark task to be the combination of 1) a publicly available dataset with a well-defined train/test split, 2) one or more clearly defined prediction tasks, and 3) pre-defined metrics for performance evaluation. We define structured data as data that is naturally represented in tabular form, and differentiate this type of data from the many other kinds in medical machine learning, including: medical images, electroencephalogram (EEG), human voice, clinical notes, biomedical text, genetics and other molecular data, and electrocardiogram (ECG) data. For each selected publication, we reviewed their proposed experimental set-up to determine if it met our definitions, and if so, we reviewed its citations to evaluate whether there was a sufficient number of results to aggregate. The data from these publications were then analyzed using R. 

\subsection{Data Availability}
The data supporting our study and conclusions will be made publicly available after the anonymous review period ends.

\section{Results}
\label{results}

The Web of Science query resulted in 1,127 publications between January 1, 1995 and September 21, 2020. It is worth noting that 1,031 (91.5\%) of these were published from 2010 onward, highlighting the trend in machine learning for healthcare over the last decade, which has also been noted by others \citep{beaulieu2019trends}. Reviewing these publications revealed many studies that have proposed benchmark datasets, and in a few cases benchmark tasks that meet our definition (see Section \ref{study selection} for this definition), across a wide array of medical disciplines. 

For example, there have been benchmark datasets proposed for peripheral blood cell recognition \citep{acevedo2020dataset}, brain tumor image segmentation \citep{menze2014multimodal}, tuberculosis identification from X-ray images \citep{jaeger2014two}, cervical cytology analysis \citep{zhang2019dccl}, glaucoma detection \citep{salam2017benchmark}, ischemic stroke lesion segmentation from MRI images \citep{maier2017isles}, seizure detection \citep{harati2014tuh}, human activity sensing and motion assessment \citep{kawaguchi2011hasc, ebert2017open}, voice disorder detection \citep{cesari2018new}, demographic trait detection from clinical notes \citep{feder2020active}, biomedical knowledge link prediction \citep{breit2020openbiolink}, molecular machine learning \citep{wu2018moleculenet}, ECG interpretation \citep{wagner2020ptb}, ICU predictions such as mortality, length of stay, patient decline, and phenotyping \citep{harutyunyan2019multitask, purushotham2018benchmarking, sheikhalishahi2020benchmarking},  neurodegenerative disorder diagnosis \citep{tagaris2018machine}, prostate cancer survival prediction \citep{guinney2017prediction},  and several tasks from the UCI machine learning repository such as predicting chronic kidney disease, diabetes, breast cancer and more \citep{Dua:2019}.

Although many benchmark datasets were found, only a small minority used structured data accompanied by publicly available train/test splits and precisely defined prediction tasks. In particular, 3 papers proposed benchmark experimental set-ups using MIMIC-III \citep{purushotham2018benchmarking, harutyunyan2019multitask, sheikhalishahi2020benchmarking}. However, the study by \citet{sheikhalishahi2020benchmarking} was only published a few months prior to this writing and had yet to be cited, leaving no results to be aggregated. The paper by \citet{purushotham2018benchmarking} had been cited approximately 80 times, but their benchmark included 6 different sets of features for training and testing, and 8 definitions of mortality prediction, making it difficult to aggregate results from other publications as there were few instances of identical overlap. 

The paper by \citet{harutyunyan2019multitask} was the one exception, having published a single train/test split of their pre-processing of the MIMIC-III dataset on Zenodo\footnote{Harutyunyan, H. et al. MIMIC-III benchmark repository. Zenodo, \url{https://doi.org/10.5281/zenodo.1306527} (2018).}, along with 4 precisely defined prediction tasks: mortality, length of stay (LOS), patient decompensation, and phenotype (ICD-9 code group). Briefly, mortality prediction was posed as a binary classification task using the first 48 hours of an ICU stay (main metric: AUROC). LOS was defined as a multi-class classification task where the LOS for each ICU stay was divided into 10 categories based on length (main metric: Cohen's kappa). Patient decompensation was defined as a series of binary classification tasks, where the outcome is death in the ensuing 24 hours and the prediction is made at every hour of an ICU stay (main metric: AUROC). Finally, phenotyping was also defined as a multi-class classification task where ICD-9 codes were divided into 25 disease groups (main metric: Macro AUROC). This study was cited approximately 190 times since its original publication in 2017, of which 19 reported on the development of one or more prediction models for at least 1 of the 4 tasks defined by this benchmark study using exactly the same experimental set-up. This resulted in a total of 210 models to compare. Among these 210 models, 172 were trained using the default data provided by the MIMIC-III benchmark, 18 used additional MIMIC-III data, and 20 used data external to MIMIC-III.

Table~\ref{studies-and-tasks} shows the tasks attempted by each of the 19 studies. It is apparent that mortality prediction is the most frequently attempted task, followed by phenotyping, while only a few studies attempted LOS prediction. Figure ~\ref{fig:performance-over-time} is a plot of the performance of each model from the papers that replicated the Harutyunyan benchmark over time. The best performing model from each paper is highlighted, and a linear model was fit to quantify the trends in performance. The slope of each regression line is provided on the figure, as well as an associated p-value. None of the slopes were significantly different from 0.

Table \ref{table:model-type-comparison} compares the performance of the most commonly used baseline models on each of the 4 prediction tasks: logistic regression, a standard LSTM RNN, a standard GRU RNN, and the Simply Attend and Diagnose (SAnD) system of \citet{song2017attend}, which is based on a novel attention mechanism. We can see that none of the deep learning models perform significantly better than logistic regression for mortality and length of stay prediction, whereas they perform slightly better on phenotype and decompensation prediction. 

In the appendix, Figure \ref{fig:AUPRC-performance} visualizes the trends in performance over time by AUPRC rather than AUROC for mortality prediction and decompensation prediction. The slopes of the linear regression are not significantly different from zero, similar to Figure \ref{fig:performance-over-time}.

\section{Discussion}
\label{discussion}

It has been noted before that a major impediment to the progress of machine learning for structured healthcare data is the lack of universal benchmarks \citep{shickel2017deep}. This has resulted in persistent debates across several fields over the relative performance of regression models (e.g. logistic regression), traditional machine learning techniques (e.g. decision trees, support vector machines, random forests), and deep learning (e.g. RNNs) \citep{christodoulou2019systematic}. 

Attempting to systematically search for benchmark experimental set-ups is challenging, particularly because of the varying usage of the term "benchmark". This term is used as both a noun (\textit{a benchmark}) and a verb (\textit{to benchmark}), making it difficult to locate benchmark experimental set-ups in the literature, as the searches are populated by many studies using the term in unrelated ways. For instance, models that are commonly used as points of comparison (e.g. logistic regression, SVM, random forest) will frequently be referred to as "benchmarks", although "baselines" would be more appropriate and would help distinguish these models from the experimental set-ups themselves in the literature. Similarly, the term is frequently used in papers that are setting "the benchmark" for a prediction task, or in other words, they are establishing the new state of the art (SOTA). And finally, the verb usage, "to benchmark", is commonly used in papers that are comparing a newly developed model to a pre-existing reference point, when the term "to compare" should suffice.

\begin{table*}[h]
  \caption{Benchmark tasks attempted by each study that used the same experimental set-up as \citet{harutyunyan2019multitask}.}
  \label{studies-and-tasks}
  \centering
  \begin{tabular}{lllll}
    \toprule
    Study     & Mortality & Length of Stay & Phenotyping & Decompensation      \\
    \midrule
    \citet{song2017attend} & & \xmark & \xmark & \xmark     \\
    \citet{bahadori2018spectral} & & & \xmark & \\
    \citet{jin2018improving} & \xmark \\
    \citet{gao2020dr} & \xmark & \xmark & \xmark & \xmark \\
    \citet{ma2020concare} & \xmark \\
    \citet{ma2020adacare} & & & \xmark & \\
    \citet{gupta2018transfer} & \xmark & & \xmark \\
    \citet{oh2019relaxed} & \xmark & \\
    \citet{bahadori2019temporal} & \xmark & \\
    \citet{gao2020stagenet} & & & & \xmark \\
    \citet{hosseini2019hierarchical} & & & \xmark & \\
    \citet{qiu2019modeling} & \xmark \\
    \citet{xu2019attention} & & & \xmark \\
    \citet{lu2019learning} & \xmark & \\
    \citet{popkes2019interpretable} & \xmark & \\
    \citet{chakraborty2019explicit} & \xmark & & \xmark & \xmark \\
    \citet{rethmeier2020efficare} & \xmark & \xmark & \xmark & \xmark \\
    \citet{sousa2020improving} & \xmark \\
    \citet{horn2019set} & \xmark \\
    \bottomrule
  \end{tabular}
\end{table*}

\begin{figure*}[h]
  \centering
  \includegraphics[width=1\linewidth]{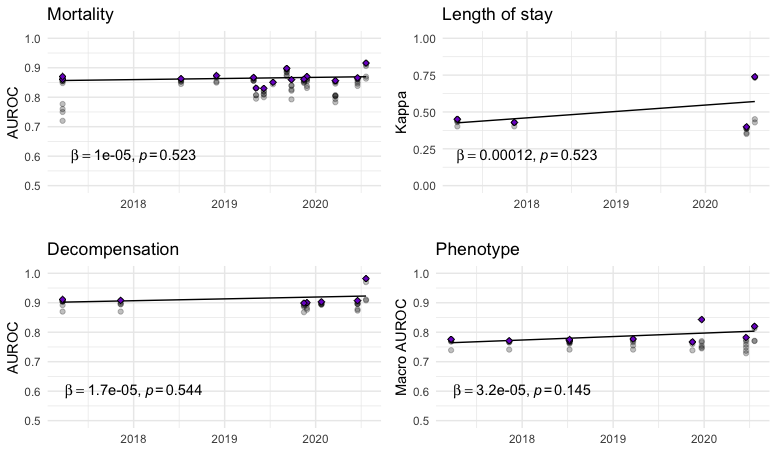}
  \caption{Performance of all models from the papers that replicated the Harutyunyan benchmark over time. The best model from each paper is indicated in purple, whereas other models are indicated in grey. Linear regression models were fit to the best model performances to quantify trends. $\beta$ is the slope of the regression and $p$ is its corresponding $p$-value. See Section \ref{results} for a definition of each of the four tasks in this figure.}
  \label{fig:performance-over-time}
\end{figure*}

\begin{table*}[h]
  \caption{Comparison of the most commonly used models to logistic regression (LR). The mean evaluation metric is provided for each model type and prediction task, with standard error in parentheses. Mortality and decompensation use AUROC, while phenotyping uses Macro AUROC and length of stay uses Cohen's kappa coefficient. Means were compared to LR using two-sided t-tests. Bolded values are significantly different from LR at the $p=0.05$ level.}
  \label{table:model-type-comparison}
  \centering
  \begin{tabular}{lllll}
    \toprule
    Prediction Task & LR & LSTM & GRU & SAnD      \\
    \hline
    Mortality & 0.834 (0.0128) & 0.850 (0.00749) & 0.847 (0.0122) & 0.851 (0.00521) \\
    Length of stay & 0.385 (0.0172) & 0.439 (0.00694) & 0.386 (*) & 0.415 (0.0138)\\
    Phenotype & 0.738 (0.00199) & \textbf{0.770 (0.000474)} & \textbf{0.767 (0.001)} & \textbf{0.768 (0.00136)} \\
    Decompensation & 0.870 (0.00113) & \textbf{0.898 (0.00348)} & \textbf{0.897 (0.00332)} & \textbf{0.895 (0.00521)}  \\
    \midrule
    \bottomrule
  \end{tabular}
  *GRU was only reported once for length of stay, so no standard error can be calculated.
\end{table*}

However, even after identifying a benchmark experimental set-up, it may still be challenging to aggregate results of separate studies in order to evaluate progress. As noted above, the paper by \citet{harutyunyan2019multitask} was cited 190 times, but only 19 were replicating the experimental set-up exactly. As one would expect, some of these citations were only referring to this benchmark to provide context for their study. However, many other studies used the experimental set-up with modifications that prevent the direct comparison of results. For example, \citet{xu2018raim} developed models for LOS and patient decompensation prediction, but used a subset of the data called the MIMIC-III Waveform Database Matched Subset. Other issues include modifying the pre-processing steps \citep{xu2018distilled}, train/test split \citep{choudhury2019differential}, task definitions \citep{islam2017marked}, applying custom patient exclusion criteria \citep{oh2018learning}, and combinations thereof. This problem is not unique to MIMIC-III benchmarks either. For example, \citet{tagaris2018machine} discusses the variability in experiments for Alzheimer's disease prediction using the ADNI dataset launched in 2008 \citep{jack2008alzheimer}.

Figure \ref{fig:performance-over-time} and Table \ref{table:model-type-comparison} further the debate over which methods are superior for this area of machine learning. For instance, there does not appear to be any significant progress in predictive performance since this benchmark was established, and logistic regression seems to do nearly as well as LSTM, GRU and attention-based models. These results raise questions about the optimal choice of datasets, pre-processing steps, task definitions, and evaluation metrics used for structured healthcare data. Each of these will be discussed in the subsequent section. 

\section{Recommendations for Future Benchmark Tasks}
Based on the results of our analysis, we offer the following concrete advice to inform future benchmarks for medical machine learning. 

\subsection{Datasets}

Unlike other areas of machine learning, the ground truth for healthcare data changes across time and space, since clinical practices are not stationary and vary between hospitals. For example, a study by \citet{nestor2019feature} demonstrated the decay in predictive performance of several model types over time using health record data. This poses a difficult challenge for deploying models in the clinic and is an important consideration in the future design of benchmark datasets and the evaluation of models. The benchmark proposed this year by \citet{sheikhalishahi2020benchmarking} used similar reasoning to argue that our selection of benchmark datasets should contain data from multiple hospitals, and models should be trained and tested on data from differing time periods. \citet{futoma2020myth} argue that the notion of generalisability in machine learning for healthcare is a myth, which would suggest that benchmarks in this field may be designed to drive either methodological progress or implementation progress, but not both.

The issue of fairness is also intertwined with our choice of datasets. For instance, if a clinical prediction model relies on features that are expensive to collect and are not routinely measured in low-to-middle-income countries (LMICs), the benefits experienced by these models will not be shared equally. \citet{deliberato2019severitas} demonstrate the process of developing a mortality prediction model using less expensive clinical variables.

\subsection{Pre-Processing Steps}

Pre-processing of longitudinal electronic healthcare data may be carried out in numerous ways, leading to many possible variations in experimental set-ups. Inclusion/exclusion criteria, handling of missing values, and handling of class imbalance are only a few examples of such steps. So long as experiments are performed on the same benchmark, these discrepancies should be minimized, although the construction of the benchmark requires making several of these decisions. A challenging, yet frequently under-estimated pre-processing step is the curation of clinical concepts necessary for the benchmark. For example, blood sugar may be measured by a bedside leukometer or by a lab test. These two measurements have different specificity and sensitivity, yet the variable for blood sugar in a dataset may combine them. In these cases, the clinical concept can only be reliably curated if meta-data on the type of measurement is present.

\subsection{Prediction Tasks}

Structured medical data, in contrast to data in computer vision, suffers from the fact that there are many possible outcome variables, and each may be of substantive interest to different groups of medical researchers. Each outcome is also high-dimensional; even a clinical concept as seemingly binary as mortality may be defined over various timespans (e.g. 30-day mortality, 1-year mortality), and using varying windows of input information (e.g. first 24 hours of an ICU stay versus first 48 hours). 

The apparent lack of progress on the Harutyunyan benchmark may be due to the choice of the tasks themselves. For instance, mortality prediction could be a poor target, since it is affected by discretionary factors such as the implicit biases of clinicians, patients and their family. A homeless person with no social support may have a different probability of dying in the ICU compared to a grandparent with family even if they have the same condition, resulting in outcome variability that is not biologically meaningful, robust or equitable.

A significant disadvantage of machine learning for structured healthcare data is that there is no proxy for Bayes error. In medical image processing, trained physicians can perform the tasks and provide a lower bound for the optimal error rate. With tasks such as mortality prediction, we cannot be sure if we are already at the Bayes error limit, and it is possible that the notion of Bayes error is not useful in this context because of the shifting ground truth of these predictions. 

\subsection{Evaluation Metrics}

In this literature review, the majority of papers reported multiple metrics, with AUROC being the most common. A potential pitfall is when papers choose non-overlapping sets of evaluation metrics, preventing direct comparisons of their results. Ideally, all benchmarks should specify a common set of evaluation metrics to be used, and all subsequent work should follow this specification. It is also important to consider how the metrics are being calculated, since different software may use different approaches. A good practice is to use open-source implementations of each evaluation metric whenever possible. A comparison of Figures \ref{fig:performance-over-time} and \ref{fig:AUPRC-performance} reveals that evaluations based on AUROC versus AUPRC arrive at similar conclusions, suggesting that they may be equally useful but providing both metrics is preferred.

\subsection{Limitations}

The review process carried out in this study may have missed other publications of benchmarks using structured medical data with enough results to aggregate and analyze. If this is the case, it is possible that new data could support a different conclusion about the state of progress in this area of machine learning for healthcare, although we believe this is unlikely. A limitation of aggregating the results on the Harutyunyan benchmark is that some studies were not aiming primarily at improving predictive performance but rather maintaining equivalent predictive performance while improving model interpretability or uncertainty estimation. Nonetheless, we thought it was necessary to aggregate these results in order to perform a comprehensive evaluation of progress on this benchmark.

\section{Conclusion}

The majority of existing benchmarks in machine learning for healthcare use medical images and far fewer exist for structured medical data. Among these, it is common that datasets are published without a full specification of an experimental set-up, such as a publicly available train/test split, precise task definitions, and evaluation metrics. Although the term "benchmark" is used very flexibly in practice, we argue that it should be reserved for referring to complete experimental set-ups to facilitate the process of evaluating progress in the field. 

After comparing the results of 20 studies that built models on the same benchmark, we found evidence that there had not been very much progress in predictive performance over time or by model type. Although strong predictive performance is not the only desirable quality of a clinical prediction model, it is nonetheless a necessary criterion. 

Nearly 20 years ago, \citet{moody2001impact} wrote an article discussing the impact of the MIT-BIH Arrhythmia Database on the automated classification of arrhythmias since it was first published in 1982 \citep{mark1982annotated}. They underscored the organizing effects the dataset had on the field and the progress it drove in ECG modelling. Today, the inter-patient accuracy of the top model is 99.5\%, \footnote{\url{https://paperswithcode.com/sota/arrhythmia-detection-on-mit-bih-ar}} suggesting that with the right alignment of our goals and practices significant improvement in modelling structured medical data should be possible.

\clearpage
\bibliography{references}

\clearpage
\appendix
\setcounter{secnumdepth}{0}
\section{Appendix}
\label{appendix}

\begin{minipage}{\textwidth}
    \centering
    \includegraphics[width=1\linewidth]{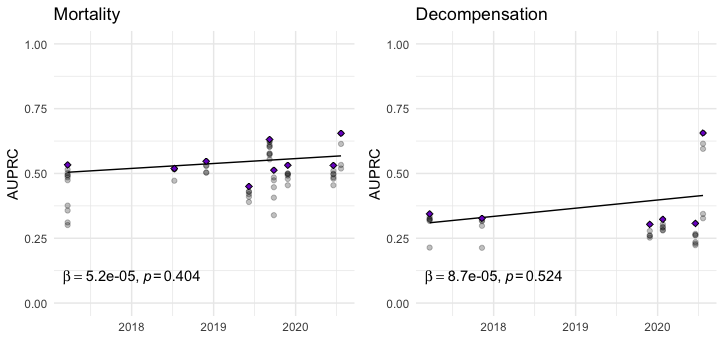}
    \captionof{figure}{Performance of models from the papers that replicated the Harutyunyan benchmark and that reported AUPRC as an evaluation metric. The best model from each paper is indicated in purple, whereas other models are indicated in grey. Linear models were fit to the best model performances to quantify trends. $\beta$ is the slope of the regression and $p$ is its corresponding $p$-value. See Section \ref{results} for a definition of mortality and decompensation prediction.}
    \label{fig:AUPRC-performance}
\end{minipage}

\end{document}